\pdfoutput=1

\documentclass[11pt]{article}

\usepackage{emnlp2021}

\usepackage{times}
\usepackage{latexsym}

\usepackage[T1]{fontenc}

\usepackage[utf8]{inputenc}

\usepackage{times}
\usepackage{latexsym}
\usepackage{todonotes}
\usepackage{tabu}
\usepackage{booktabs}
\usepackage{amsfonts, bm, amsmath}
\usepackage{multirow, tabularx}
\usepackage{times}
\usepackage{latexsym}
\usepackage[T1]{fontenc}
\usepackage[utf8]{inputenc}
\usepackage{microtype}
\usepackage{arydshln}
\usepackage[capitalise]{cleveref}
\usepackage{caption}
\usepackage{subcaption}
\usepackage{threeparttable}

\def\bb{\mathbf{b}}

\def\bv{\mathbf{v}}
\def\bw{\mathbf{w}}

\def\cD{\mathcal{D}}
\def\cE{\mathcal{E}}

\def\cG{\mathcal{G}}

\def\cN{\mathcal{N}}

\def\cV{\mathcal{V}}

\DeclareMathOperator*{\roberta}{RoBERTa}
\DeclareMathOperator*{\gal}{GAL}
\title{Contrastive Document Representation Learning \\
with Graph Attention Networks}

\author{Peng Xu\quad Xinchi Chen\quad Xiaofei Ma\quad Zhiheng Huang\quad Bing Xiang \\ \texttt{\{pengx, xcc, xiaofeim, zhiheng, bxiang\}@amazon.com} \\ \\ AWS AI Labs}

\begin{document}
\maketitle
\begin{abstract}
Recent progress in pretrained Transformer-based language models has shown great success in learning contextual representation of text. However, due to the quadratic self-attention complexity, most of the pretrained Transformers models can only handle relatively short text. It is still a challenge when it comes to modeling very long documents. In this work, we propose to use a graph attention network on top of the available pretrained Transformers model to learn document embeddings. This graph attention network allows us to leverage the high-level semantic structure of the document. In addition, based on our graph document model, we design a simple contrastive learning strategy to pretrain our models on a large amount of unlabeled corpus. Empirically, we demonstrate the effectiveness of our approaches in document classification and document retrieval tasks. 
\end{abstract}

\section{Introduction}
Document representations that capture the semantics are crucial to various document-level Natural Language Processing (NLP) tasks, including sentiment analysis \cite{medhat2014sentiment}, text classification \cite{kowsari2019text} and information retrieval \cite{lin2020pretrained}.
In recent years, an increasing volume of work has focused on learning a task-agnostic universal representation for long documents. While improved performance in downstream tasks have been achieved, there are two challenges towards learning a high quality document representation: (1) \textbf{absence of document structure}. Most works treat the document as a sequence of tokens without considering high-level structure. (2) \textbf{data scarcity}. Existing methods in document representation learning are significantly affected by the scarcity of document-level data. 

Transformers-based pretrained language models are ubiquitously state-of-the-art across many NLP tasks. Transformer models such as BERT \cite{devlin-etal-2019-bert} and its variants have shown great success in learning contextual representation of text. Representation from large language models can partially mitigate the data scarcity issue due to pretraining on a large amounts of unlabeled data. 
However, those models mostly consider token-level information and their pretraining tasks are not directly targeting long document representations. Another issue of directly applying transformer-based models is the limit of the input text length. 
Due to the quadratic complexity of self-attention, most of the pretrained transformers models can only handle a relatively short text.
A wide spectrum of efficient, fast transformer models (collectively called ``X-formers'') have been proposed to tackle this problem; e.g., Longformer \cite{beltagy2020longformer} and Bigbird \cite{zaheer2020big} use sparse attention to improve the computational and memory efficiency for long sequence text. Nevertheless, these models still focus on token-level interactions without considering high-level semantic structure of the document. 

Recently, there is a resurgence of interest in Contrastive Learning (CL) due to its success in self-supervised representation learning in computer vision \cite{chen2020simple,he2020momentum}. 
Contrastive Learning offers a simple method to learn disentangled representation that encodes invariance to small and local changes in the input data without using any labeled data.  
In NLP domain, contrastive learning has been employed to learn sentence
representation \cite{wu2020clear,qu2020coda} under either self-supervised or supervised settings.

\begin{figure}[t!]
  \centering\includegraphics[width=0.8\linewidth]{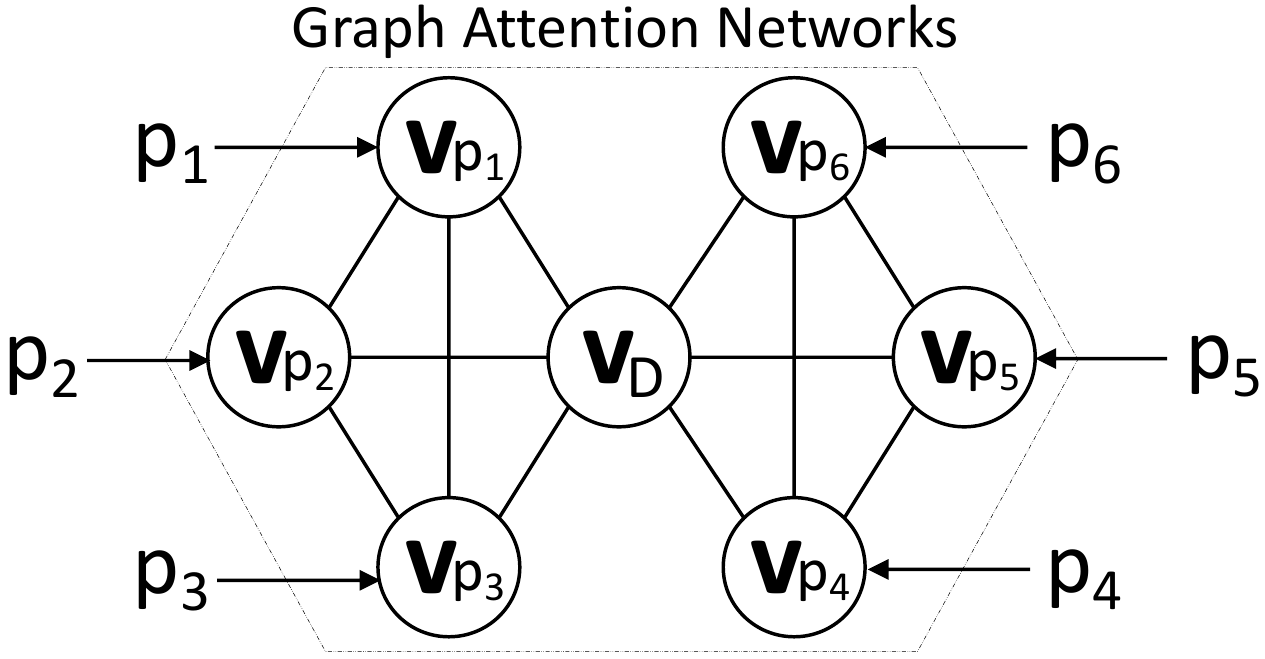}
  \caption{An example of the Graph-Roberta architecture for the document representation.}  \label{fig:graphdoc}
\end{figure}

In this work, we propose a Graph Attention Network (GAT) based model that explicitly utilizes the high-level semantic structure of the documents to learn document embeddings. 
 We model the document as not just a sequence of text, but a collection of passages or sentences. Specifically, the proposed model introduces a graph on top of the document passages (Fig. \ref{fig:graphdoc}) to utilize multi-granularity information. First, passages are encoded using RoBERTa \cite{liu2019roberta} to collect word-level knowledge. Then passages are connected to leverage the higher-level structured information. At last, a graph attention network \cite{velivckovic2017graph} is applied to obtain the multi-granularity document representation. To better learn the document embedding, we propose a document-level contrastive learning strategy to pretrain our models. In our contrastive learning framework, we split the document into random sub-documents and train the model to maximize the agreement over the representations of the sub-documents that come from the same document. This simple strategy allows us to pretrain our models on a large unlabelled corpus without any additional priors. As we will see, this simple pretraining task indeed helps the model on the downstream tasks.


The contributions of this paper can be summarized as follows.
\begin{itemize}
    \item We propose a graph document model with graph attention networks that can not only explicitly utilize the high-level structure of the document but also leverage pretrained Transformer encoders to obtain low-level contextual information.
    \item We propose a simple document-level contrastive learning strategy, which does not require any handcrafted transformations and is suitable for large-scale pretraining. 
    \item We conduct empirical evaluations on our models and contrastive pretraining strategy. We show that our graph-roberta models achieve great performance on both document classification and retrieval tasks. Specifically we demonstrate that our contrastive pretraining helps the model learn a meaningful document representation even without fine-tuning, and improve both the training convergence speed and final performance during end-to-end finetuning on downstream classification tasks. For document retrieval tasks, we demonstrate that our graph-roberta models have great semantic matching performance, compensating the typical lexical matching system.
        
\end{itemize}
\section{Methodology}

In this section, we describe our main model and contrastive pretraining strategy. 

        
\subsection{Graph Document Architecture}
In this work, we model a document as graph over passages. Given a document $D$ with passages $\{p_1, \ldots, p_{|D|}\}$, we define an undirected graph $\cG:(\cV,\cE)$, where $\cV$ consists of $n+1$ nodes $(v_D, v_{p_1}, \cdots, v_{p_n})$ and the graph edges $\cE$ are constructed based on the document structure. An example of a document graph is shown in \cref{fig:graphdoc}. Once the document graph is defined, we can instantiate a neural network model based on the graph structure. 
\paragraph{Passage Node Initialization} 
First, we use the state-of-the-art contextual language models to encode each passage text, since each passage is relatively short. Specifically, given a passage $p_i$ consists of a sequence of words $\{w_{i,1}, w_{i,2}, \cdots, w_{i,|p_i|}\}$, we use Roberta\cite{liu2019roberta} as the encoder model for the passage node and project the \texttt{[CLS]} vector into fixed embedding space as the initial passage node representation. 
\begin{equation}
    \bv_{p_i}^{(0)} = \tanh (\bw \phi(w_{i,\cdot}) + \bb), \label{eq:init}
\end{equation}
where $\phi$ is $\roberta$ with \texttt{[CLS]} vector. 
\paragraph{Document Node Initialization}
For the document node, we simply use the average of all the passage node embeddings as the initial representation.
\begin{equation}
    \bv_{D}^{(0)} = \frac{1}{n}\sum_{i=1}^{n} \bv_{p_i}^{(0)}.
\end{equation}
\paragraph{Graph Attention Layers}
Finally, we apply $T$ Graph Attention Layers (GAL)\footnote{Refer to \citet{velivckovic2017graph} for the details of GAL.} to aggregate all the information from different nodes.
\begin{equation}
    \bv_{i}^{(t+1)} = \gal(\bv_{k}^{(t)}| k \in \cN(i)),
    \label{eq:gan}
\end{equation}
where $\cN(i)$ is the neighbour node set of passage node $p_i$ on the given graph structure. The step $t$ counts from 1 to $T$ and the final document node representation is $\bv_{D}^{(T)}$. 

\subsection{Contrastive Pretraining for Document Representation Learning}
\begin{figure}[t!]
  \centering\includegraphics[width=0.85\linewidth]{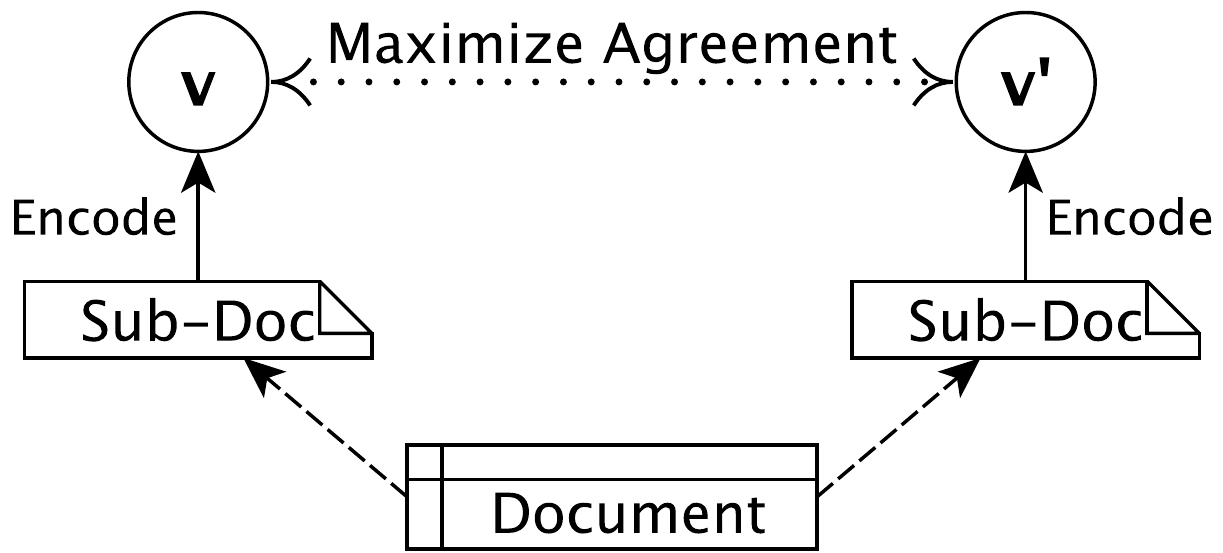}
  \caption{A simple contrastive learning strategy for document representation.}  \label{fig:cl}
\end{figure}

We design a simple contrastive learning task to pretrain our graph document models. The main idea follows from the contrastive learning framework in \cite{chen2020simple}, where the task is to learn an encoder function to maximize the agreement between augmented views of the same image. Here we consider that any proportions inside the same document are the different ``views''. The task is to maximize the agreement between different proportions that come from the same document. Since our document has been represented as a list of passages, a proportion of a document would be any subset of passages, which we call a \textit{sub-document}.  During training time, we randomly sample a mini-batch of $N$ documents $\cD = \{D_i\}_{i=1}^N$. 
For each document $D_i$, we randomly split passages to two subsets as sub-documents:
\begin{equation}
    D_i \xrightarrow{} \tilde{D}_i, \tilde{D}^\prime_i,
\end{equation}
where $D_i$ is the union set of $\tilde{D}_i$ and $\tilde{D}^\prime_i$.

We treat $\tilde{D}_i$ and $\tilde{D}^\prime_i$ as the positive pair. Any pair of sub-documents that come from different documents are negative pairs. Then the noise contrastive loss function for a positive pair is defined as 
\begin{equation}
\begin{aligned}
    \ell (\tilde{D}_i) &= -\log \dfrac{\exp(\bv_{\tilde{D}_i} \cdot \bv_{\tilde{D}^\prime_i})}{\sum_{j=1}^N \exp(\bv_{\tilde{D}_i} \cdot \bv_{\tilde{D}^\prime_j})}, \\
    \ell (\tilde{D}^\prime_i) &= -\log \dfrac{\exp(\bv_{\tilde{D}_i} \cdot \bv_{\tilde{D}^\prime_i})}{\sum_{j=1}^N \exp(\bv_{\tilde{D}_j} \cdot \bv_{\tilde{D}^\prime_i})},
\end{aligned}
\label{eq:loss}
\end{equation}
where $\bv_{\tilde{D}_i}$ is the encoding of the sub-document $\tilde{D}_i$ based on the proposed graph document model. 

The final loss is computed across all the pairs.
\begin{equation}
\ell(\cD) = \dfrac{1}{2N} \sum_{i=1}^N \ell(\tilde{D}_i) + \ell(\tilde{D}^\prime_i).
\label{eq:overall_loss}
\end{equation}

\section{Experiments}

We experiment on two popular applications, text classification and document retrieval to evaluate the proposed approach. The experimental results show that the graph based document representation could capture long document information and the contrastive learning strategy could utilize unlabeled data to further improve the performance and training efficiency.

\paragraph{Model details} Throughout the paper, we use roberta-\textit{base} model \cite{liu2019roberta} as our passage node encoder. On top of that, we add 2 graph attention layers with 2 heads and skip connections. Specifically, we utilize the Deep Graph Library\footnote{https://github.com/dmlc/dgl} for GAT implementation. The embedding size for passage and document nodes is 512. We refer this model as graph-roberta model.

\subsection{Datasets}
In this section, we describe all the datasets we use in this paper. 

\textbf{OpenWebText} \cite{Gokaslan2019OpenWeb} is an open-source recreation of the Webtext corpus in \citet{radford2019language}. The text was extracted from Reddit post urls, which produces around 8M documents.

\textbf{arXiv} \cite{he2019long} is a collection of 33,388 arXiv scientific papers from 11 categories. The average document length exceeds 5,000 words. We create a random train/dev/test split of 25,568/3,196/3,197. 

\textbf{Newsgroup} \cite{lang1995newsweeder}  a collection of newsgroup documents, partitioned (nearly) evenly across 20 different newsgroups. It contains 11,314 training and 7,532 test samples. We sample 10\% of the training data for validation.

\textbf{IMDB} \cite{maas2011learning} is a dataset for binary sentiment classification. It contains 25,000 labeled movie reviews as the training set and another 25,000 movie reviews as the test set. We random sample 1,000 examples from the training set for validation.

\textbf{Hyperpartisan} \cite{kiesel2019semeval} is a binary classification dataset for hyperpartisan news detection. It consists of 645 documents in total. We use the same train/dev/test split (516/64/65) from \cite{beltagy2020longformer}.

\textbf{Robust04} \cite{voorhees2005} is the news collection from the TREC 2004 Robust track. It is a document retrieval dataset consisting of 249 queries with relevance labels on a corpus of 528K documents. 

\textbf{MSMARCO DR} \cite{bajaj2016ms} is a document ranking dataset with about 3.2M documents. It provides over 367K training queries and an official dev set of 5,193 queries. The Trec 2019 Deep Learning track \cite{craswell2020overview} also provides an additional test set of 43 queries.

\textbf{WIKIR}\footnote{https://github.com/getalp/wikIR} \cite{frej2019wikir} is an open-source toolkit to create large-scale information retrieval datasets based on Wikipedia. In this work, we use the English Wikipedia dump from 2020/12/20\footnote{https://archive.org/download/enwiki-20201220/enwiki-20201220-pages-articles-multistream.xml.bz2} and following the the same settings in \citet{frej2019wikir} except for that we preserve the punctuation and section information in the document. We obtain two datasets, WIKIR62K and WIKIRS62K, both of which contain around 60k training queries, 1k dev queries and 1k test queries. The queries in WIKIR62K are built based on titles and the ones in WIKIRS62K are based on the first sentences. The processed document corpus size is around 2.4M documents. 

Since graph-roberta models take a document input as a graph over passages, we split each document into passages with around 100 words while respecting the sentence boundary. For WIKIR documents, we also respect the section boundaries. Without additional specification, we use the fully-connected graph structure by default.

\subsection{Document Classification}
In this section, we conduct empirical evaluation of our models on document classification tasks. We consider 4 datasets, arXiv, Hyperpartisan, IMDB and Newsgroup.
We compare our graph-roberta models with the baseline model Roberta \cite{liu2019roberta}, as well as Longformer \cite{beltagy2020longformer} and BigBird \cite{zaheer2020big}, two state-of-the-art transformer models that handle long text input\footnote{We take the \texttt{[CLS]} embedding from those models as the document representation.}. In our experiments, we only consider the \textit{base} version of those models.

\paragraph{Contrastive Pretraining} We pretrain our graph-roberta models on \texttt{OpenWebText} dataset. During the training process of contrastive learning, for each document, we keep up to 50 passages and we randomly select half the number of passages as the sub-document and the rest of the half as the other sub-document. We train for 10 epochs with batch size 1,536, using Adam \cite{kingma2014adam} optimizer with a learning rate 5e-5 and warm up rate 0.1. 

\paragraph{Finetuning} For graph-roberta models, we keep up to 50 passages per document during training and at inference time, we keep up to 100 passages per document. For the other models, we truncate the document text up to the maximum sequence length they are allowed to handle; Roberta's maximum input length is 512, and Longformer and BigBird's maximum input length is 4,096. The detailed training configurations are shown in appendix.
\begin{figure*}[htbp]
  \centering\includegraphics[width=0.9\linewidth]{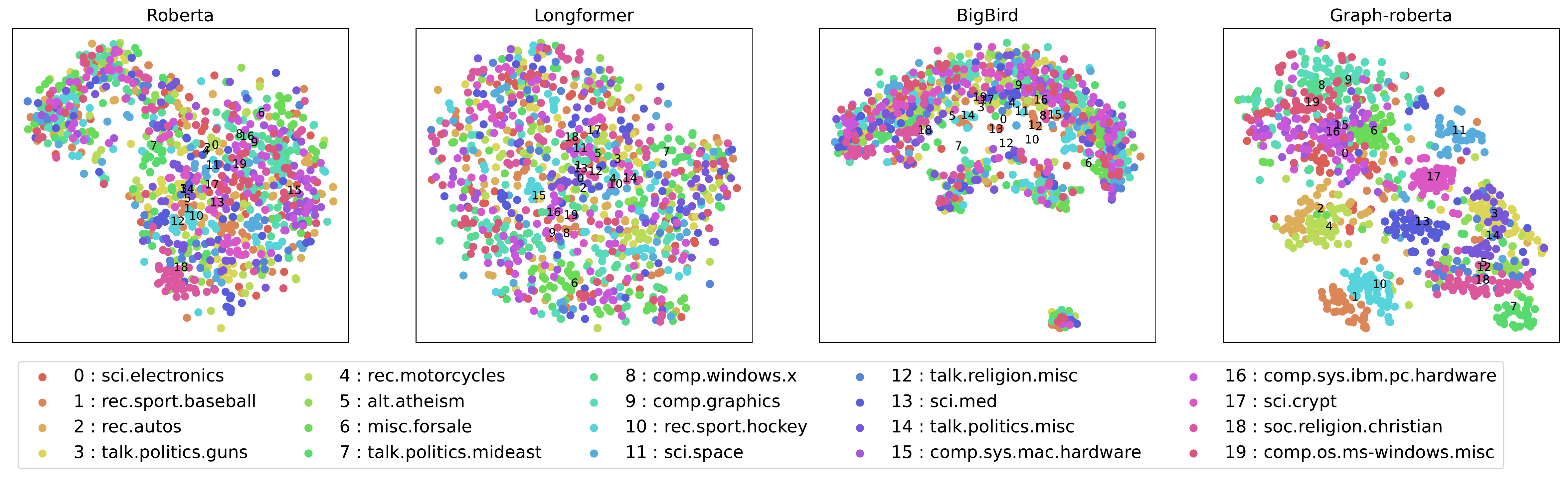}
  \caption{TSNE visualization of different representations on 1000 documents sampled from Newsgroup dataset.}  \label{fig:tsne}
\end{figure*}
\begin{figure*}[htbp]
  \centering\includegraphics[width=0.9\linewidth]{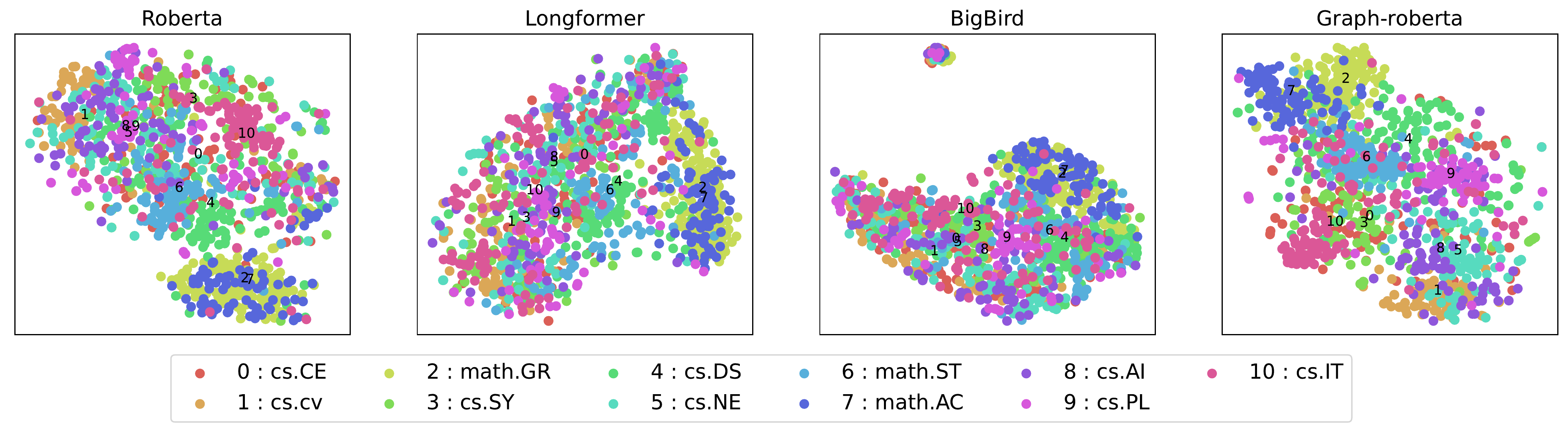}
  \caption{TSNE visualization of different representations on 1000 documents sampled from arXiv dataset.}  \label{fig:tsne2}
\end{figure*}

\paragraph{Clustering} 
First, we evaluate the capability of our graph-roberta model as an off-the-shelf document encoder through document clustering. 
We take the document node representation with the pretrained graph-roberta model and \texttt{[CLS]} embeddings from the other three models. We run k-means clustering methods on the training set and run inference on the test set. We compute the normalized mutual information (NMI) and Purity to evaluate the clustering quality. We report the results on arXiv and Newsgroup dataset in \cref{tab:cluster}. As we can see, our pretrained graph-roberta model clearly outperforms the other three models by a large margin. This is expected that the other three models are not pretrained on any document-level tasks. \cref{fig:tsne} \& \ref{fig:tsne2} showcase that the simple unsupervised contrastive learning strategy indeed helps graph-roberta model learn meaningful document representations.
\begin{table}[ht]
\centering
\footnotesize
\begin{tabular}{lcccc}
\toprule
\multirow{2}{*}{\textbf{model}} & \multicolumn{2}{c}{\textbf{arXiv}} & \multicolumn{2}{c}{\textbf{Newsgroup}} \\
 & NMI & Purity & NMI & Purity \\
 \midrule
Roberta & 0.267 & 0.327 & 0.116 & 0.160 \\
Longformer & 0.168 & 0.241 & 0.084 & 0.133 \\
Bigbird & 0.180 & 0.261 & 0.059 & 0.119 \\
Graph-roberta & \textbf{0.437} & \textbf{0.558} & \textbf{0.516} & \textbf{0.475} \\
\bottomrule
\end{tabular}
\caption{Clustering performance with different document embeddings on the test sets.}
\label{tab:cluster}
\end{table}

\begin{figure}[ht!]
  \centering\includegraphics[width=0.49\linewidth]{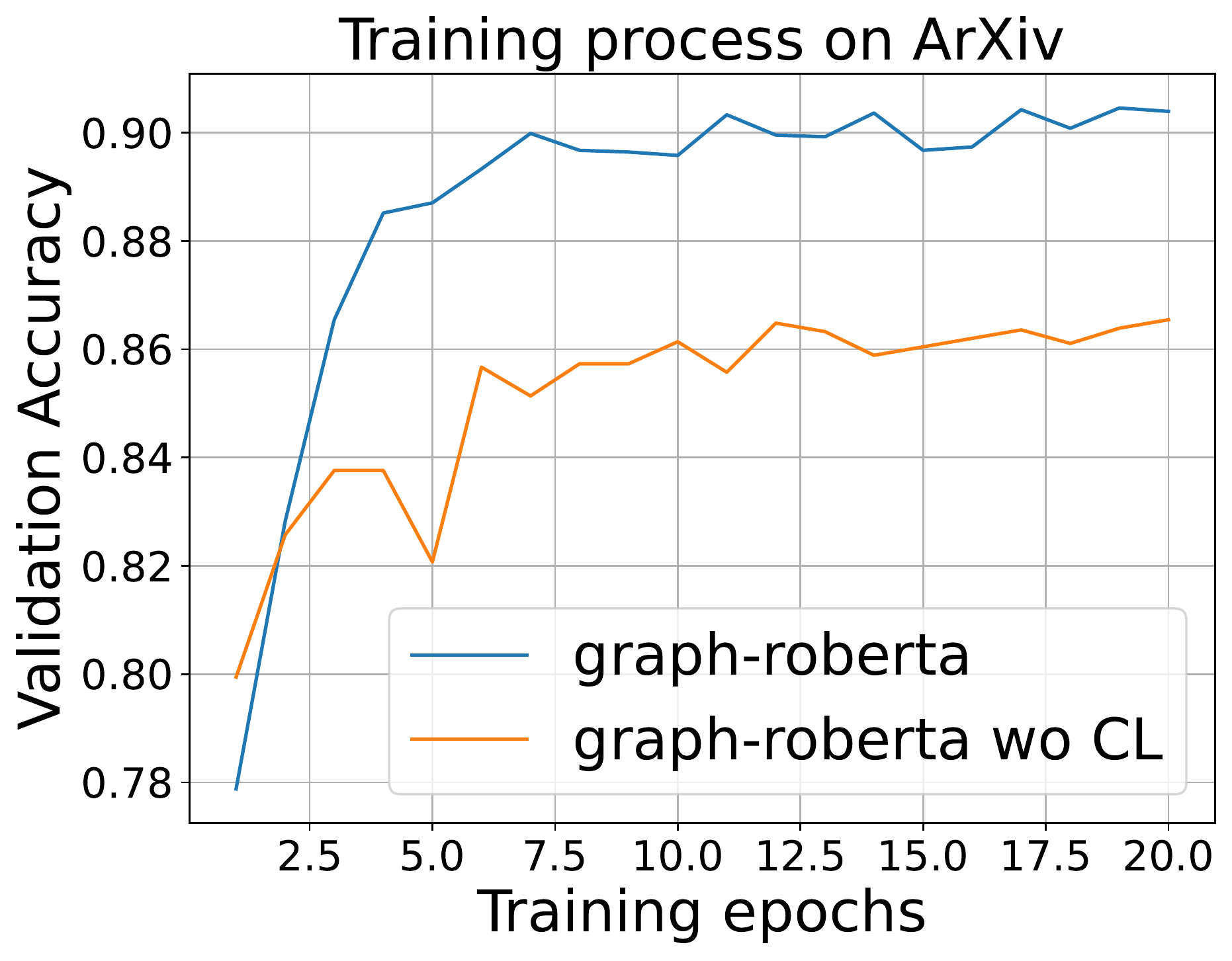}
  \centering\includegraphics[width=0.49\linewidth]{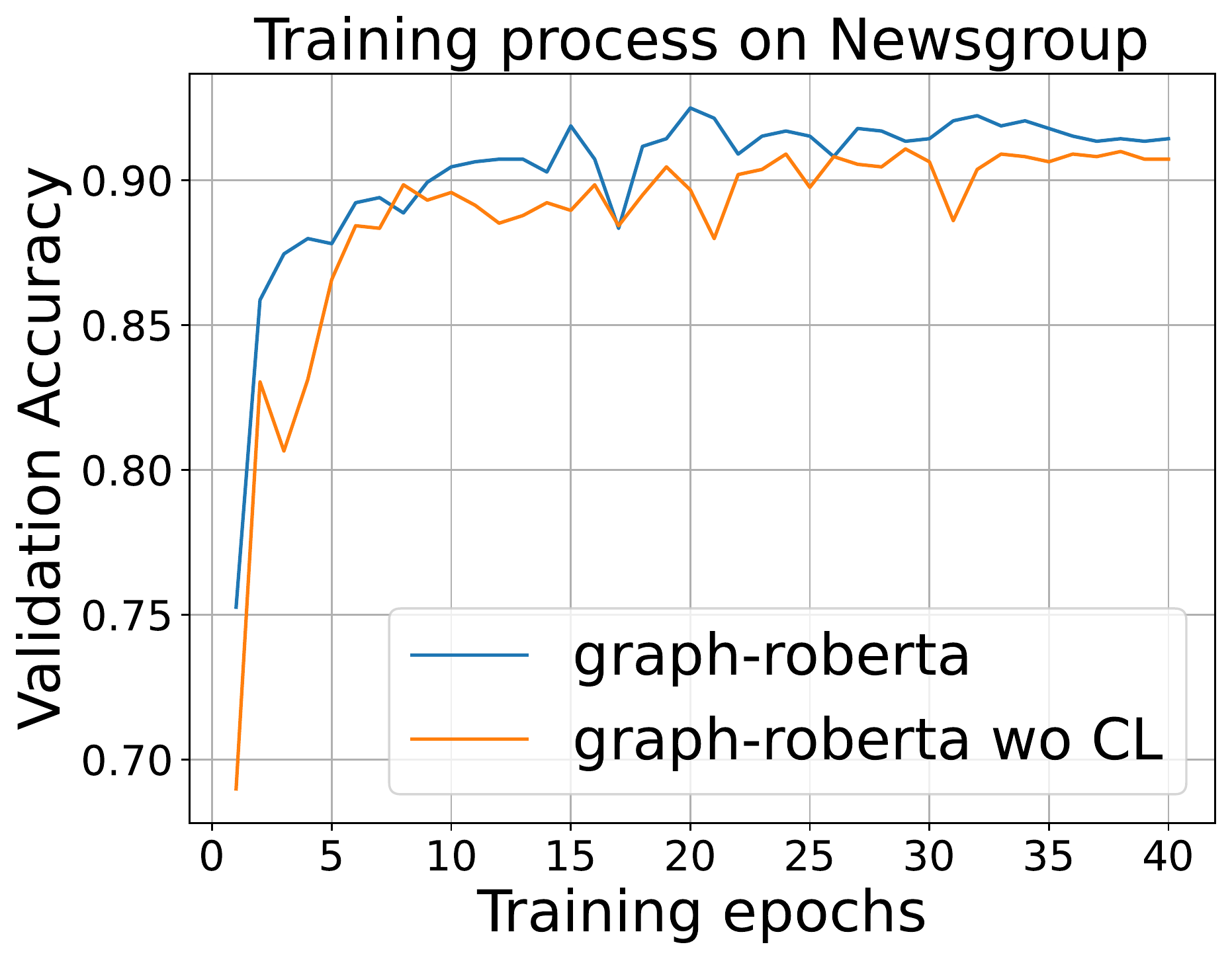}
  \caption{Validation accuracy over the training process.}  \label{fig:training}
\end{figure}


\begin{table*}[!h]
\centering
\footnotesize
\begin{tabular}{lccccc}
\toprule
\textbf{Model} & \textbf{arXiv} & \textbf{Newgroup} & \textbf{IMDB} & \textbf{Hyperpartisan} & \textbf{\textit{avg}} \\
\midrule
Roberta & 89.99   & 86.22   & 95.63   & 90.77  & 90.65 \\
Longformer & 90.90   & 86.42   &\textbf{95.77} & 92.31  &   91.35\\
BigBird & 88.99   & 81.41   & 95.32   & 89.23  &   88.74\\
Graph-roberta w/o CL & 86.83   & 85.36   & 94.32   & 94.12  &  90.16\\
\textbf{Graph-roberta} & \textbf{91.21} & \textbf{86.66} & 94.26   & \textbf{96.15}  & \textbf{92.07} \\ 
\midrule
\midrule
\footnotemark[6]Longformer \cite{beltagy2020longformer} & - & -  &  95.70 &94.80&-\\
\footnotemark[6]BigBird \cite{zaheer2020big} &92.31& -  & 95.20  &92.20&-\\
\bottomrule
\end{tabular}
\caption{End-to-end classification performance of different models on the test set. The numbers are in percent.}
\label{tab:dc-end2end}
\end{table*}
\addtocounter{footnote}{1}
\vspace{-5mm}
\paragraph{End-to-end Classification} To evaluate the full capability of the graph-roberta model, we also conduct end-to-end finetuning on the 4 datasets. In addition to 4 pretrained models, we also report the performance of graph-roberta without contrastive learning. The results are shown in \cref{tab:dc-end2end}. First, we can see that graph-roberta model outperforms all the other methods on 3 out of 4 datasets. The exception is IMDB dataset, which has relatively short text. Also we see that contrastive learning indeed helps improve the final performance. 
\cref{fig:training} shows the end-to-end  training processes of graph-roberta models on arXiv and Newsgroup datasets. It demonstrates that contrastive learning task speeds up the finetuning progress and helps learn a better model. 
\footnotetext[6]{The reported numbers from \citet{beltagy2020longformer,zaheer2020big} on arXiv and Hyperpartisan datasets are not comparable with ours because they did not release the train/test split of the data.}

\subsection{Document Retrieval}

In this section, we extend our model to embedding-based document retrieval task. In this case, we consider the query as a single-node graph, where the representation is computed by the initial node representation. With that, we apply dot-product similarity to retrieve relevant documents. Our approach is essentially representation-based model.

\paragraph{Contrastive Pretraining} To better align with the retrieval tasks, during pretraining, we sample one passage from each document as the sub-document and the rest as the other sub-document instead of an even random split, and we only compute the contrastive loss over the long sub-documents. In addition, we set 50\% of the time we select the first passage of each document and 50\% of the time we sample uniformly from the document. This is very similar to the Inverse Cloze Task (ICT) introduced from \citet{lee2019latent}, except for the difference that ICT randomly selects one sentence from the passage whereas we randomly selects passages from the document. We pretrain the graph-roberta model on the \texttt{OpenWebText} data for 10 epochs with batch size 1,024, using Adam \cite{kingma2014adam} optimizer with learning rate 5e-5 and warmup rate 0.1. 

\paragraph{Finetuning} To finetune the model on ranking datasets, we use the similar training loss in the contrastive pretraining except we use the actual training queries-documents pairs. Besides in-batch negatives, we also sample additional negative candidates either uniformly or from some hard negatives such as the top BM25 retrieval pool for each training query.

First, we run the experiment with Robust04 dataset. We train and cross-validate machine learning models on the given 5 folds. During each run, we finetune the model 10 epochs with batch size 32, using Adam optimizer with learning rate 5e-5 and warmup rate 0.1. We sample additional 8 random negatives uniformly for each training query. We compare our models with BM25 baseline. The results are shown in \cref{tab:robust04}. First we can see that the contrastive pretraining significantly improves the graph-roberta model performance (e.g. P@20 improves by over 100\%). Still as a retrieval model, graph-roberta underperforms BM25. We conjecture that there are two reasons. (1) robust04 query set is too small to train such a complex neural representation model. (2) robust04 queries are all short key word queries, which favor lexical-matching methods such as BM25 over contextual transformer models. Nevertheless, we combine the retrieval results from graph-roberta model and BM25 through a weighted average of their scores (the weight is selected through cross-validation), we improve the nDCG@20 by 2\% in absolute value over BM25, which indicates our model compensates BM25 results for semantic matching. 
\begin{table}[ht]
\centering
\footnotesize
\begin{tabular}{lcc}
\toprule
\multirow{2}{*}{\textbf{Model}} & \multicolumn{2}{c}{\textbf{Robust04}}  \\
     & \multicolumn{1}{c}{\textbf{nDCG@20}} & \multicolumn{1}{c}{\textbf{P@20}} \\
\midrule
BM25  & 41.63  & 35.68  \\
\midrule
Graph-roberta w/o CL  & 13.97    & 11.39        \\
Graph-roberta        & 20.02    & 23.89        \\
Graph-roberta +BM25        & \textbf{43.90}    & \textbf{37.21}\\
\bottomrule
\end{tabular}
\caption{Document retrieval on the test set of Robust04 dataset. The numbers are in percent.}
\label{tab:robust04}
\end{table}

Now we present the experiment on much larger document ranking dataset MSMARCO. 
We finetune graph-roberta models on MSMARCO training set with batch size 128 for 10 epochs. For each training query, we also sample one hard negative in addition to the batch negatives. For the first 5 epochs, we randomly sample one negative from the top 100 BM25 retrieval results. For the latter 5 epochs, we randomly sample one from the top 100 results retrieved using the 5-epoch checkpoint model. We also sample 100 queries from the official training set as our own validation set to monitor the training progress. We report the retrieval performance (without reranking) in \cref{tab:ms}. 

\begin{table}[hb]
\centering
\footnotesize
\begin{tabular}{lcc}
\toprule
\multirow{2}{*}{\textbf{Model}} & \multicolumn{1}{c}{\textbf{Dev}} & \multicolumn{1}{c}{\textbf{Trec DL}} \\
     & \multicolumn{1}{c}{\textbf{MRR}}   & \multicolumn{1}{c}{\textbf{nDCG@10}} \\
\midrule
BM25    & 25.87  & 52.97    \\
DE-Hybrid-E \cite{luan2020sparse}   & 28.70   & 59.50     \\
ME-Hybrid-E \cite{luan2020sparse}   & 31.00   & 61.00     \\
ACNE \textit{FirstP} \cite{xiong2020approximate}   & 37.40   & \textbf{61.50}    \\
\midrule
Graph-roberta w/o CL  & 33.69  & 50.95    \\
Graph-roberta    & 34.85  & 54.05    \\
Graph-roberta +BM25    & \textbf{37.60}  & 61.44     \\
\bottomrule
\end{tabular}
\caption{Document retrieval on the test set of MSMARCO document dataset. The numbers are in percent.}
\label{tab:ms}
\end{table}

\begin{table*}[!ht]
\centering
\footnotesize
\begin{tabular}{llccccccc}
\toprule
\multirow{2}{*}{\textbf{Model}} & \multirow{2}{*}{\textbf{Graph}} & \multicolumn{3}{c}{\textbf{WIKIR62K (Title)}} & \multicolumn{3}{c}{\textbf{WIKIRS62K (First Sentence)}} \\
\cmidrule(r){3-5} \cmidrule(r){6-8}
 &  & \textbf{P@5} & \textbf{P@10} & \textbf{P@20} & \textbf{P@5} & \textbf{P@10} & \textbf{P@20} \\
\midrule
BM25     & -       & 28.66  & 22.17  & 16.62   & 22.20    & 16.74   & 12.65            \\
MatchPyramid \cite{pang2016text} & -       & 20.76  & 16.83   & 13.11   &  20.74  & 16.76  & 12.89                                  \\
ConvKNRM \cite{dai2018convolutional} & -       &    17.46  & 14.87  & 12.16  & 18.94 & 16.51 & 13.05   \\
\midrule
Graph-roberta w/o CL & full & 20.68 & 16.66 & 12.86 & 20.92 & 17.39 & 13.51 \\
Graph-roberta & full & 22.98 & 18.46 & 14.19 & 24.02 & 19.65 & 15.39 \\
Graph-roberta + BM25 & full & 39.38 & 30.52 & 21.98 & 35.24 & \textbf{27.53} & 19.74 \\
\midrule
Graph-roberta w/o CL & section & 20.16 & 16.07 & 12.56 & 21.98 & 17.93 & 13.86 \\
Graph-roberta & section & 23.32 & 18.69 & 14.16 & 23.78 & 19.78 & 15.27 \\
Graph-roberta + BM25 & section & \textbf{39.70} & \textbf{30.64} & \textbf{22.13} & \textbf{35.40} & 27.45 & \textbf{19.86} \\
\bottomrule
\end{tabular}
\caption{Document retrieval benchmark on the test sets of WIKIR datasets. The numbers are in percent.}
\label{tab: wikir}
\end{table*}

Similarly to Robust04 experiment, contrastive learning as a pretraining strategy again improves the graph-roberta model performance. Note the improvement on MSMARCO is not as significant as in Robust04. Considering the fact that MSMARCO has a much larger training set, it is expected that the benefit of pretraining is less. Comparing with BM25, graph-roberta as a dense retrieval method achieves almost 9 points better in MRR@100 on the official dev set. We also list the performance of the state-of-the-art neural retrieval methods. DE-Hybrid-E and ME-Hybrid-E methods \cite{luan2020sparse} are the two hybrid sparse-dense models that combining BM25 and BERT encoded dense presentations. Note that graph-roberta already outperforms the hybrid models on the official dev set, indicating that the representation learned by the graph-roberta model is very effective. Lastly, combining graph-roberta and BM25 retrieval results through simple weighted average, gives us the similar performance by the SOTA method ACNE
\footnote{ACNE \textit{MaxP} which produces the best numbers (MRR 38.38\% on the official dev set) is not based on learned document embeddings, but on a set of passage representations for each document.} \cite{xiong2020approximate}. Furthermore, we believe that the training strategy introduced in ACNE can also be applied on graph-roberta model training and we leave it to the future work.

To further demonstrate the effectiveness of graph-roberta models for document retrieval task, we evaluate our models on the two large document retrieval datasets created via WIKIR \cite{frej2019wikir}, namely WIKIR62K and WIKIRS62K.  
In this experiment, we also consider different graph structures in modeling the Wikipedia documents. Besides the default fully-connected graph, we also consider the section structure information in the documents. Specifically, we consider the structure that all the passages within each section are mutually connected. The document node and the first passage nodes are connected with each other. We denote this graph as the \textit{section} graph. We finetune the models on the training data for 5 epochs with batch size 128, using Adam optimizer with learning rate 2e-5 and warmup rate 0.1. For each query, we also sample one hard negative from the top 100 BM25 retrieved candidates.
The final retrieval benchmark is shown in \cref{tab: wikir}.

In \cref{tab: wikir}, we see that contrastive pretraining consistently helps improve the model performance in both title queries and first-sentence queries. 
BM25 performs much better for title queries than the first-sentence queries, as observed in \citet{frej2019wikir} since title queries are usually keyword queries. Our graph-roberta model outperforms MatchPyramid \cite{pang2016text} and ConvKRNM \cite{dai2018convolutional}, and performs consistently on both title and first-sentence queries. We further combine the results of graph-roberta and BM25. Overall, the ensemble of BM25 and graph-roberta gives the best results. 

\begin{figure}[htb]
    \centering
    \begin{subfigure}[b]{0.33\textwidth}
    \includegraphics[width=\textwidth]{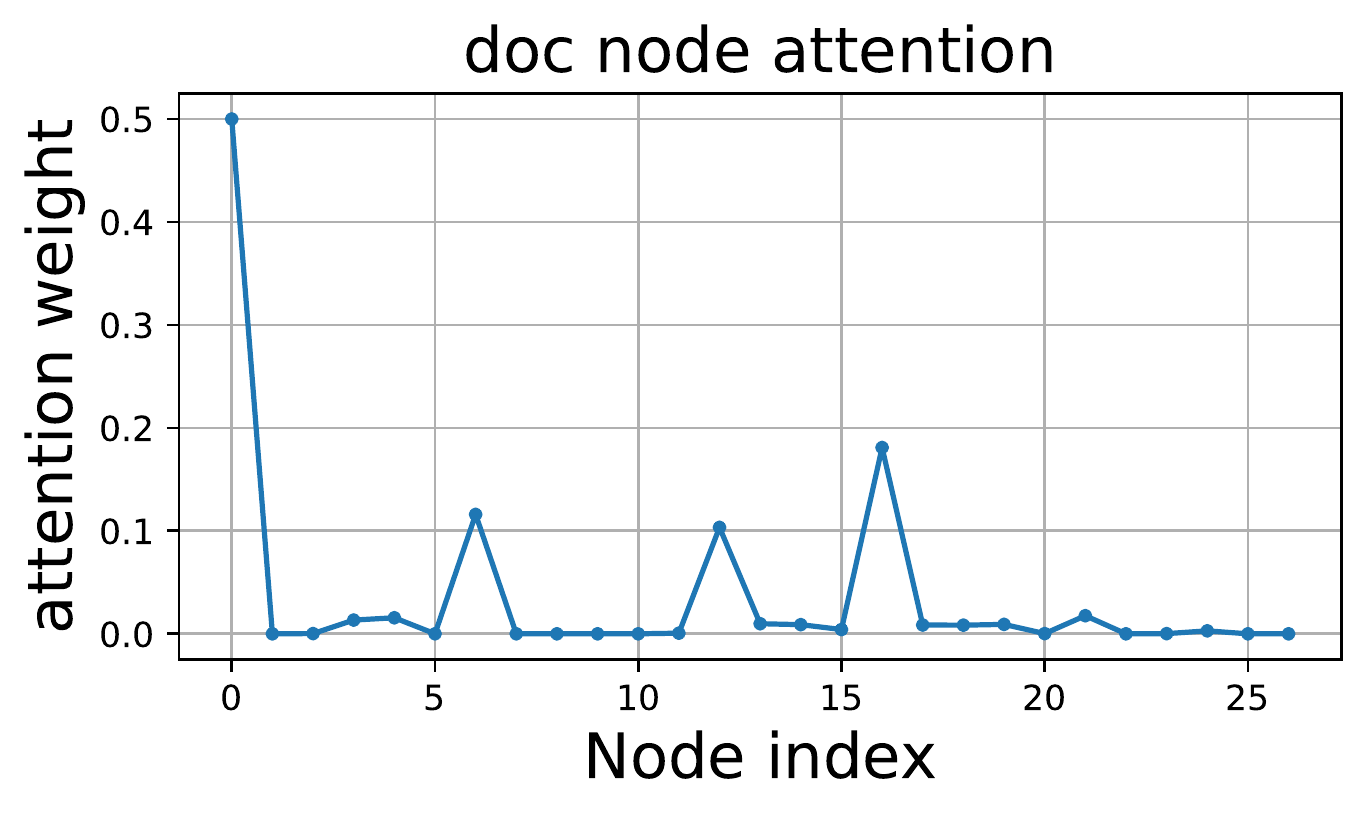}
    \caption{Fully-connected graph}
    \label{fig:docattention}
    \end{subfigure}
    \begin{subfigure}[b]{0.33\textwidth}
    \includegraphics[width=\textwidth]{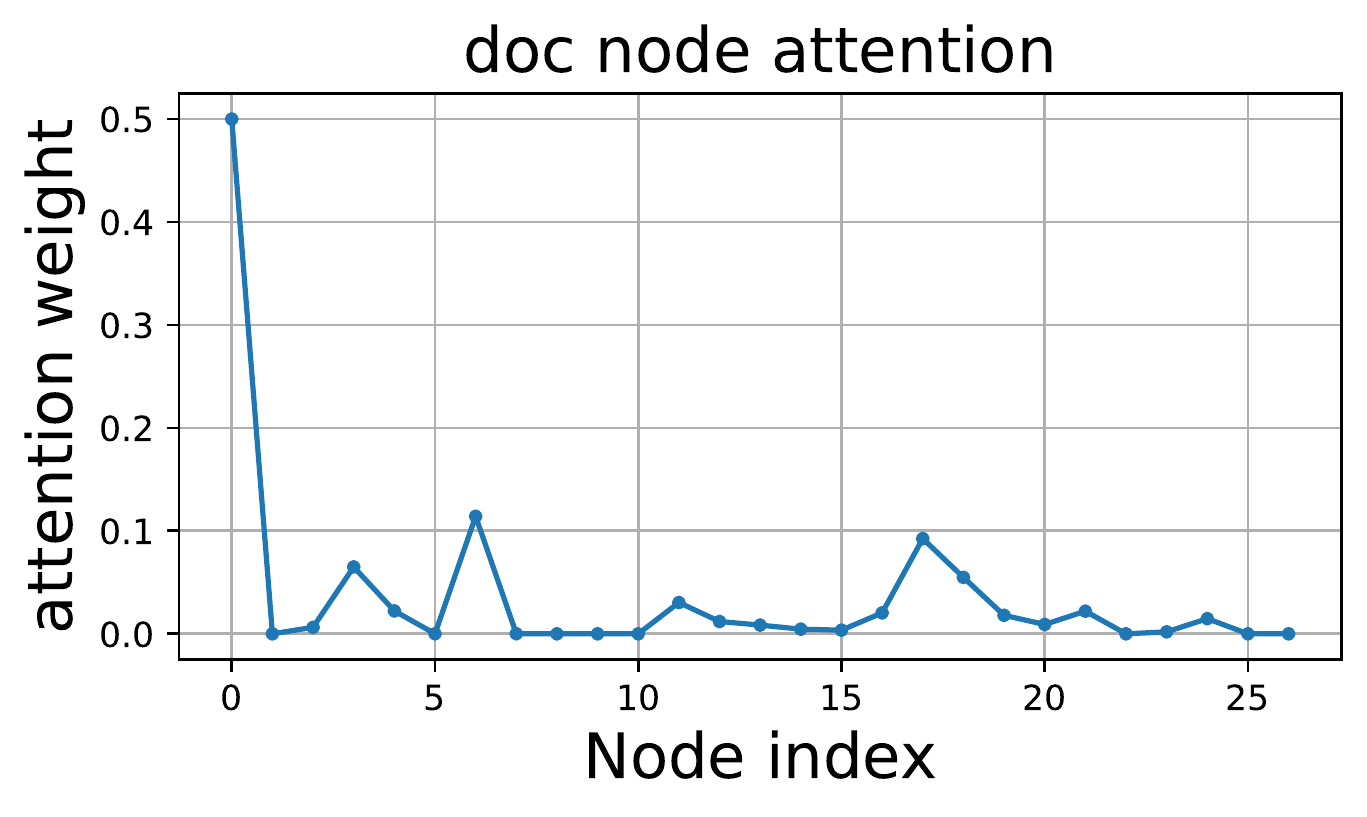}
    \caption{section graph}
    \label{fig:docattention}
    \end{subfigure}
    \caption{An example of the document node attention patterns on graph-roberta models. The axis is the graph node index. Node index 0 is the document node and the rest are passage nodes.}
    \label{fig:docatt}
\end{figure}

\begin{figure}[htb]
    \centering
    \begin{subfigure}[b]{0.235\textwidth}
    \includegraphics[width=\textwidth]{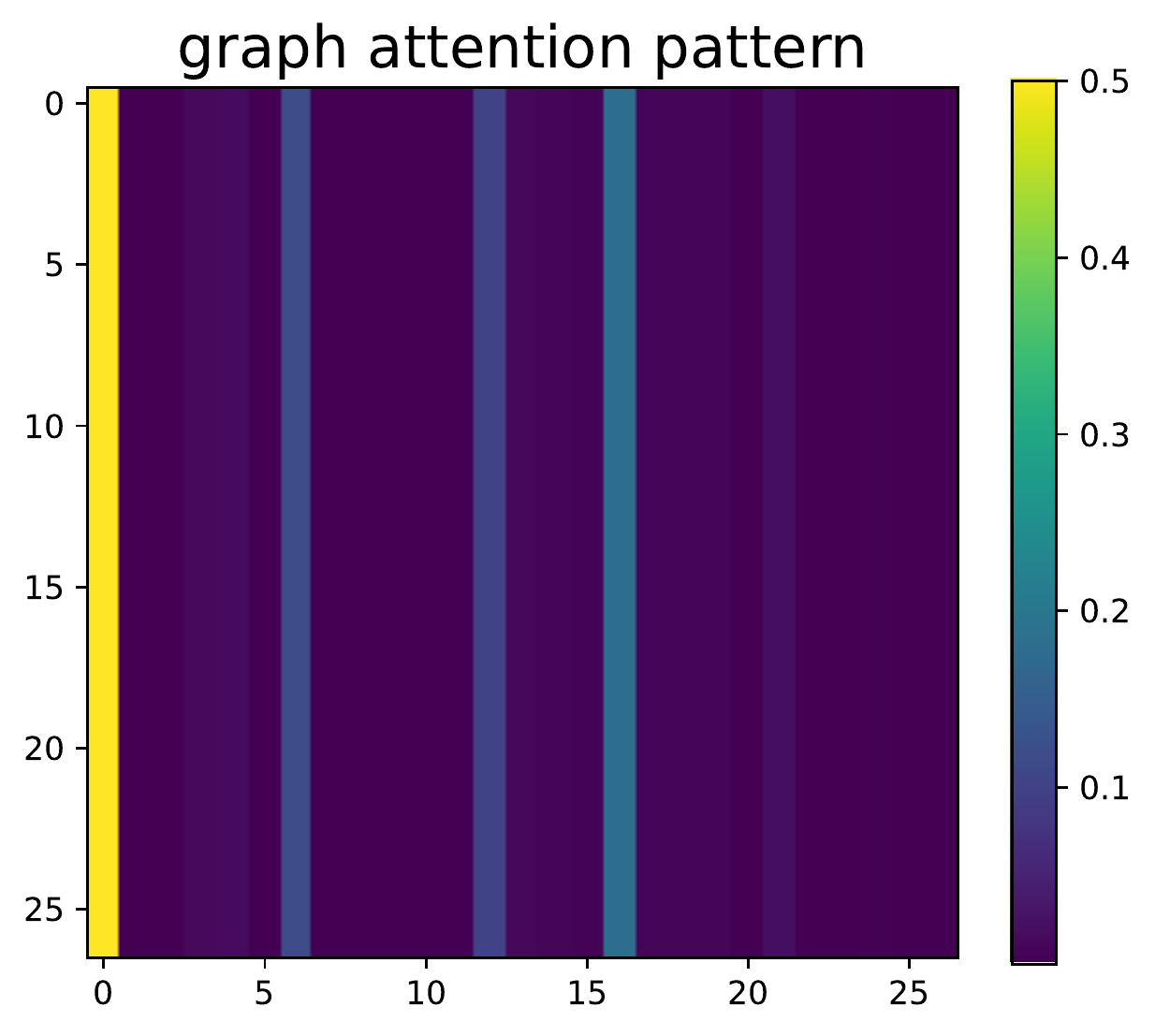}
    \caption{fully-connected graph}
    \label{fig:docattention}
    \end{subfigure}
    \begin{subfigure}[b]{0.235\textwidth}
    \includegraphics[width=\textwidth]{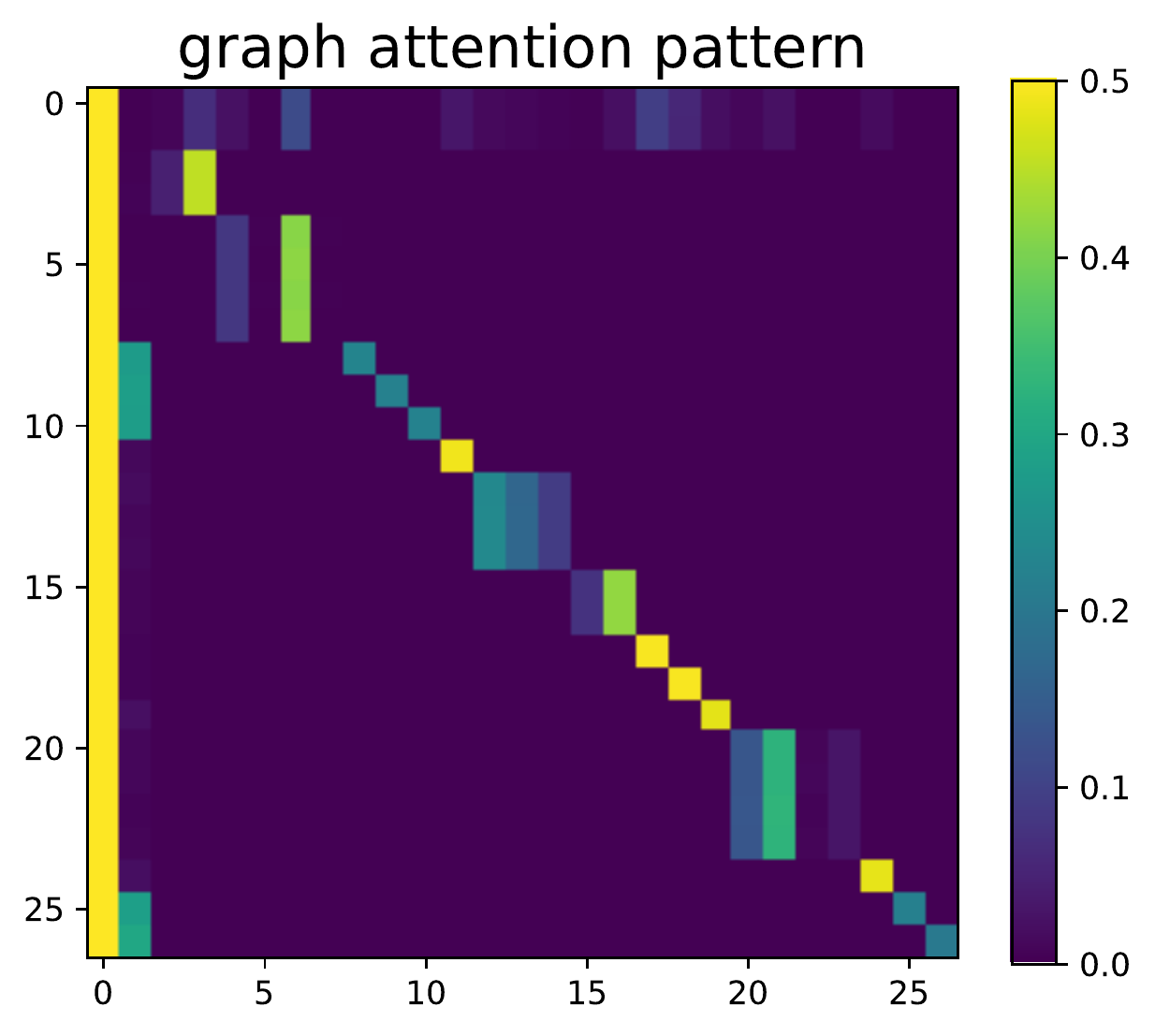}
    \caption{section graph}
    \label{fig:docattention}
    \end{subfigure}
    \caption{An example of the graph attention pattern on graph-roberta models. The axes are graph node index. Node index 0 is the document node and the rest are passage nodes.}
    \label{fig:gatt}
\end{figure}

We notice that for graph-roberta models, utilizing the section graph as described earlier performs slightly better than the default fully-connected graph, although the difference is small. We conjecture that on this dataset, the document representation does not rely much on the interaction between passages. 
We look into the graph attention patterns by the two models (graph-roberta with fully-connected graph and graph-roberta with section graph). We compute the average attention weights of the last graph attention layer. We observe that on both models, the document node usually attends to similar passages. As an example, we plot the graph attention weights on both models. \cref{fig:docatt} shows the attention weights of the document node. As we can see, for this example, both models attend to similar passages besides the document node. In \cref{fig:gatt}, we observe that for graph-roberta with the fully-connected graph, all the other passage nodes have similar attention patterns as the document node, while for graph-roberta with the section graph, the passage nodes can actually learn some nontrivial patterns, which we believe could be beneficial for more complex tasks.

\section{Related work}

\paragraph{Document Representation Learning}
One line of related work is to utilize the successful pretrained Transformer models \cite{radford2018improving,devlin-etal-2019-bert} to obtain contextual text representations. It has been shown to be successful on sentences and short passages in textual similarity tasks and passage retrieval tasks \cite{reimers2019sentence,minaee2021deep,karpukhin-etal-2020-dense,liang2020embedding}.
\citet{he2019long,dai-etal-2019-transformer,zaheer2020big} propose sparse attention to enable the transformer models to handle long text sequences more efficiently. However these pretrained models still focus on token-level interactions. 
\citet{jiang2019semantic} proposed a hierarchical attention model on top of recurrent neural network to tackle text matching of long documents, which later was extended by \citet{yang2020beyond} to transformer architectures. Both works only focus on the text matching task. 
\citet{pappagari2019hierarchical} proposed a hierarchical transformer model to encode long documents, where they apply a recurrent network or transformer layer on top of the original BERT model. In our work, we use a GAT network which can better leverage the existing document structure and we design a simple and effective contrastive learning framework based on our graphic model.

Another line of related work is to use graph neural network for document modeling. \citet{peng2018large,peng2019hierarchical} proposed to use graph convolutional networks (GCN) to model document as a graph of words, which allows the model to capture long-distance semantics. \citet{yao2019graph} built a single graph for a whole corpus based on both word-to-word and document-to-word relations, which is learned by a GCN model.

\paragraph{Contrastive Learning}
Contrastive learning used as a self-supervised pretraining method has been widely used in NLP models \cite{rethmeier2021primer}. Token or sentence-level contrastive learning tasks have been shown to be very useful in learning better contextual presentations
\cite{clark-etal-2020-pre, giorgi2020declutr,meng2021coco}.
There also have been works that propose data augmentations for contrastive learning.
\citet{fang2020cert} proposed to use back-translation to construct positive sentence pairs in their contrastive learning framework.
\citet{wu2020clear,qu2020coda} proposed multiple sentence-level augmentations strategies to do sentence contrastive learning.
Most of these work still focus on either local token-level tasks or short sentence-level tasks. In our work, we directly work on document-level contrastive learning task. More recently \citet{luo2021unsupervised} proposed to use multiple data augmentations such as synonym substitution and back-translation to do unsupervised document representation learning. The difference in our work is that we have a much simpler framework that does not require those hand-craft transformations and we demonstrate that our contrastive learning strategy as a pretraining task can help the downstream tasks across various datasets.

\section{Conclusions}
In this work, we propose a simple graph attention network model to learn document embeddings. Our model not only can leverage the recent advancement of pretrained Transformer models as building blocks, but also explicitly utilize the high-level structure of the documents. In addition, we propose a simple document-level contrastive learning strategy that does not require handcraft transformations. With this strategy, we conduct large scale contrastive pretraining on a large corpus. Empirically we demonstrate our methods achieve great performance on both document classification and document retrieval tasks. 


\bibliography{anthology,custom}

\begin{thebibliography}{46}
\expandafter\ifx\csname natexlab\endcsname\relax\def\natexlab#1{#1}\fi

\bibitem[{Bajaj et~al.(2016)Bajaj, Campos, Craswell, Deng, Gao, Liu, Majumder,
  McNamara, Mitra, Nguyen et~al.}]{bajaj2016ms}
Payal Bajaj, Daniel Campos, Nick Craswell, Li~Deng, Jianfeng Gao, Xiaodong Liu,
  Rangan Majumder, Andrew McNamara, Bhaskar Mitra, Tri Nguyen, et~al. 2016.
\newblock Ms marco: A human generated machine reading comprehension dataset.
\newblock \emph{arXiv preprint arXiv:1611.09268}.

\bibitem[{Beltagy et~al.(2020)Beltagy, Peters, and
  Cohan}]{beltagy2020longformer}
Iz~Beltagy, Matthew~E Peters, and Arman Cohan. 2020.
\newblock Longformer: The long-document transformer.
\newblock \emph{arXiv preprint arXiv:2004.05150}.

\bibitem[{Chen et~al.(2020)Chen, Kornblith, Norouzi, and
  Hinton}]{chen2020simple}
Ting Chen, Simon Kornblith, Mohammad Norouzi, and Geoffrey Hinton. 2020.
\newblock A simple framework for contrastive learning of visual
  representations.
\newblock In \emph{International conference on machine learning}, pages
  1597--1607. PMLR.

\bibitem[{Clark et~al.(2020)Clark, Luong, Le, and
  Manning}]{clark-etal-2020-pre}
Kevin Clark, Minh-Thang Luong, Quoc Le, and Christopher~D. Manning. 2020.
\newblock \href {https://doi.org/10.18653/v1/2020.emnlp-main.20} {Pre-training
  transformers as energy-based cloze models}.
\newblock In \emph{Proceedings of the 2020 Conference on Empirical Methods in
  Natural Language Processing (EMNLP)}, pages 285--294, Online. Association for
  Computational Linguistics.

\bibitem[{Craswell et~al.(2020)Craswell, Mitra, Yilmaz, Campos, and
  Voorhees}]{craswell2020overview}
Nick Craswell, Bhaskar Mitra, Emine Yilmaz, Daniel Campos, and Ellen~M
  Voorhees. 2020.
\newblock Overview of the trec 2019 deep learning track.
\newblock \emph{arXiv preprint arXiv:2003.07820}.

\bibitem[{Dai et~al.(2018)Dai, Xiong, Callan, and Liu}]{dai2018convolutional}
Zhuyun Dai, Chenyan Xiong, Jamie Callan, and Zhiyuan Liu. 2018.
\newblock Convolutional neural networks for soft-matching n-grams in ad-hoc
  search.
\newblock In \emph{Proceedings of the eleventh ACM international conference on
  web search and data mining}, pages 126--134.

\bibitem[{Dai et~al.(2019)Dai, Yang, Yang, Carbonell, Le, and
  Salakhutdinov}]{dai-etal-2019-transformer}
Zihang Dai, Zhilin Yang, Yiming Yang, Jaime Carbonell, Quoc Le, and Ruslan
  Salakhutdinov. 2019.
\newblock \href {https://doi.org/10.18653/v1/P19-1285} {Transformer-{XL}:
  Attentive language models beyond a fixed-length context}.
\newblock In \emph{Proceedings of the 57th Annual Meeting of the Association
  for Computational Linguistics}, pages 2978--2988, Florence, Italy.
  Association for Computational Linguistics.

\bibitem[{Devlin et~al.(2019)Devlin, Chang, Lee, and
  Toutanova}]{devlin-etal-2019-bert}
Jacob Devlin, Ming-Wei Chang, Kenton Lee, and Kristina Toutanova. 2019.
\newblock \href {https://doi.org/10.18653/v1/N19-1423} {{BERT}: Pre-training of
  deep bidirectional transformers for language understanding}.
\newblock In \emph{Proceedings of the 2019 Conference of the North {A}merican
  Chapter of the Association for Computational Linguistics: Human Language
  Technologies, Volume 1 (Long and Short Papers)}, pages 4171--4186,
  Minneapolis, Minnesota. Association for Computational Linguistics.

\bibitem[{Fang et~al.(2020)Fang, Wang, Zhou, Ding, and Xie}]{fang2020cert}
Hongchao Fang, Sicheng Wang, Meng Zhou, Jiayuan Ding, and Pengtao Xie. 2020.
\newblock Cert: Contrastive self-supervised learning for language
  understanding.
\newblock \emph{arXiv preprint arXiv:2005.12766}.

\bibitem[{Frej et~al.(2019)Frej, Schwab, and Chevallet}]{frej2019wikir}
Jibril Frej, Didier Schwab, and Jean-Pierre Chevallet. 2019.
\newblock Wikir: A python toolkit for building a large-scale wikipedia-based
  english information retrieval dataset.
\newblock \emph{arXiv preprint arXiv:1912.01901}.

\bibitem[{Giorgi et~al.(2020)Giorgi, Nitski, Bader, and
  Wang}]{giorgi2020declutr}
John~M Giorgi, Osvald Nitski, Gary~D Bader, and Bo~Wang. 2020.
\newblock Declutr: Deep contrastive learning for unsupervised textual
  representations.
\newblock \emph{arXiv preprint arXiv:2006.03659}.

\bibitem[{Gokaslan et~al.(2019)Gokaslan, Cohen, Pavlick, and
  Tellex}]{Gokaslan2019OpenWeb}
Aaron Gokaslan, Vanya Cohen, Ellie Pavlick, and Stefanie Tellex. 2019.
\newblock Openwebtext corpus.
\newblock \url{http://Skylion007.github.io/OpenWebTextCorpus}.

\bibitem[{He et~al.(2019)He, Wang, Liu, Feng, and Wu}]{he2019long}
Jun He, Liqun Wang, Liu Liu, Jiao Feng, and Hao Wu. 2019.
\newblock Long document classification from local word glimpses via recurrent
  attention learning.
\newblock \emph{IEEE Access}, 7:40707--40718.

\bibitem[{He et~al.(2020)He, Fan, Wu, Xie, and Girshick}]{he2020momentum}
Kaiming He, Haoqi Fan, Yuxin Wu, Saining Xie, and Ross Girshick. 2020.
\newblock Momentum contrast for unsupervised visual representation learning.
\newblock In \emph{Proceedings of the IEEE/CVF Conference on Computer Vision
  and Pattern Recognition}, pages 9729--9738.

\bibitem[{Jiang et~al.(2019)Jiang, Zhang, Li, Bendersky, Golbandi, and
  Najork}]{jiang2019semantic}
Jyun-Yu Jiang, Mingyang Zhang, Cheng Li, Michael Bendersky, Nadav Golbandi, and
  Marc Najork. 2019.
\newblock Semantic text matching for long-form documents.
\newblock In \emph{The World Wide Web Conference}, pages 795--806.

\bibitem[{Karpukhin et~al.(2020)Karpukhin, Oguz, Min, Lewis, Wu, Edunov, Chen,
  and Yih}]{karpukhin-etal-2020-dense}
Vladimir Karpukhin, Barlas Oguz, Sewon Min, Patrick Lewis, Ledell Wu, Sergey
  Edunov, Danqi Chen, and Wen-tau Yih. 2020.
\newblock \href {https://doi.org/10.18653/v1/2020.emnlp-main.550} {Dense
  passage retrieval for open-domain question answering}.
\newblock In \emph{Proceedings of the 2020 Conference on Empirical Methods in
  Natural Language Processing (EMNLP)}, pages 6769--6781, Online. Association
  for Computational Linguistics.

\bibitem[{Kiesel et~al.(2019)Kiesel, Mestre, Shukla, Vincent, Adineh, Corney,
  Stein, and Potthast}]{kiesel2019semeval}
Johannes Kiesel, Maria Mestre, Rishabh Shukla, Emmanuel Vincent, Payam Adineh,
  David Corney, Benno Stein, and Martin Potthast. 2019.
\newblock Semeval-2019 task 4: Hyperpartisan news detection.
\newblock In \emph{Proceedings of the 13th International Workshop on Semantic
  Evaluation}, pages 829--839.

\bibitem[{Kingma and Ba(2014)}]{kingma2014adam}
Diederik~P Kingma and Jimmy Ba. 2014.
\newblock Adam: A method for stochastic optimization.
\newblock \emph{arXiv preprint arXiv:1412.6980}.

\bibitem[{Kowsari et~al.(2019)Kowsari, Jafari~Meimandi, Heidarysafa, Mendu,
  Barnes, and Brown}]{kowsari2019text}
Kamran Kowsari, Kiana Jafari~Meimandi, Mojtaba Heidarysafa, Sanjana Mendu,
  Laura Barnes, and Donald Brown. 2019.
\newblock Text classification algorithms: A survey.
\newblock \emph{Information}, 10(4):150.

\bibitem[{Lang(1995)}]{lang1995newsweeder}
Ken Lang. 1995.
\newblock Newsweeder: Learning to filter netnews.
\newblock In \emph{Machine Learning Proceedings 1995}, pages 331--339.
  Elsevier.

\bibitem[{Lee et~al.(2019)Lee, Chang, and Toutanova}]{lee2019latent}
Kenton Lee, Ming-Wei Chang, and Kristina Toutanova. 2019.
\newblock Latent retrieval for weakly supervised open domain question
  answering.
\newblock \emph{arXiv preprint arXiv:1906.00300}.

\bibitem[{Liang et~al.(2020)Liang, Xu, Shakeri, Santos, Nallapati, Huang, and
  Xiang}]{liang2020embedding}
Davis Liang, Peng Xu, Siamak Shakeri, Cicero Nogueira~dos Santos, Ramesh
  Nallapati, Zhiheng Huang, and Bing Xiang. 2020.
\newblock Embedding-based zero-shot retrieval through query generation.
\newblock \emph{arXiv preprint arXiv:2009.10270}.

\bibitem[{Lin et~al.(2020)Lin, Nogueira, and Yates}]{lin2020pretrained}
Jimmy Lin, Rodrigo Nogueira, and Andrew Yates. 2020.
\newblock Pretrained transformers for text ranking: Bert and beyond.
\newblock \emph{arXiv preprint arXiv:2010.06467}.

\bibitem[{Liu et~al.(2019)Liu, Ott, Goyal, Du, Joshi, Chen, Levy, Lewis,
  Zettlemoyer, and Stoyanov}]{liu2019roberta}
Yinhan Liu, Myle Ott, Naman Goyal, Jingfei Du, Mandar Joshi, Danqi Chen, Omer
  Levy, Mike Lewis, Luke Zettlemoyer, and Veselin Stoyanov. 2019.
\newblock Roberta: A robustly optimized bert pretraining approach.
\newblock \emph{arXiv preprint arXiv:1907.11692}.

\bibitem[{Luan et~al.(2020)Luan, Eisenstein, Toutanova, and
  Collins}]{luan2020sparse}
Yi~Luan, Jacob Eisenstein, Kristina Toutanova, and Michael Collins. 2020.
\newblock Sparse, dense, and attentional representations for text retrieval.
\newblock \emph{arXiv preprint arXiv:2005.00181}.

\bibitem[{Luo et~al.(2021)Luo, Cheng, Ni, Yu, Zhang, Zong, Liu, Chen, Song,
  Chen et~al.}]{luo2021unsupervised}
Dongsheng Luo, Wei Cheng, Jingchao Ni, Wenchao Yu, Xuchao Zhang, Bo~Zong,
  Yanchi Liu, Zhengzhang Chen, Dongjin Song, Haifeng Chen, et~al. 2021.
\newblock Unsupervised document embedding via contrastive augmentation.
\newblock \emph{arXiv preprint arXiv:2103.14542}.

\bibitem[{Maas et~al.(2011)Maas, Daly, Pham, Huang, Ng, and
  Potts}]{maas2011learning}
Andrew Maas, Raymond~E Daly, Peter~T Pham, Dan Huang, Andrew~Y Ng, and
  Christopher Potts. 2011.
\newblock Learning word vectors for sentiment analysis.
\newblock In \emph{Proceedings of the 49th annual meeting of the association
  for computational linguistics: Human language technologies}, pages 142--150.

\bibitem[{Medhat et~al.(2014)Medhat, Hassan, and Korashy}]{medhat2014sentiment}
Walaa Medhat, Ahmed Hassan, and Hoda Korashy. 2014.
\newblock Sentiment analysis algorithms and applications: A survey.
\newblock \emph{Ain Shams engineering journal}, 5(4):1093--1113.

\bibitem[{Meng et~al.(2021)Meng, Xiong, Bajaj, Tiwary, Bennett, Han, and
  Song}]{meng2021coco}
Yu~Meng, Chenyan Xiong, Payal Bajaj, Saurabh Tiwary, Paul Bennett, Jiawei Han,
  and Xia Song. 2021.
\newblock Coco-lm: Correcting and contrasting text sequences for language model
  pretraining.
\newblock \emph{arXiv preprint arXiv:2102.08473}.

\bibitem[{Minaee et~al.(2021)Minaee, Kalchbrenner, Cambria, Nikzad, Chenaghlu,
  and Gao}]{minaee2021deep}
Shervin Minaee, Nal Kalchbrenner, Erik Cambria, Narjes Nikzad, Meysam
  Chenaghlu, and Jianfeng Gao. 2021.
\newblock Deep learning--based text classification: A comprehensive review.
\newblock \emph{ACM Computing Surveys (CSUR)}, 54(3):1--40.

\bibitem[{Pang et~al.(2016)Pang, Lan, Guo, Xu, Wan, and Cheng}]{pang2016text}
Liang Pang, Yanyan Lan, Jiafeng Guo, Jun Xu, Shengxian Wan, and Xueqi Cheng.
  2016.
\newblock Text matching as image recognition.
\newblock In \emph{Proceedings of the AAAI Conference on Artificial
  Intelligence}, volume~30.

\bibitem[{Pappagari et~al.(2019)Pappagari, Zelasko, Villalba, Carmiel, and
  Dehak}]{pappagari2019hierarchical}
Raghavendra Pappagari, Piotr Zelasko, Jes{\'u}s Villalba, Yishay Carmiel, and
  Najim Dehak. 2019.
\newblock Hierarchical transformers for long document classification.
\newblock In \emph{2019 IEEE Automatic Speech Recognition and Understanding
  Workshop (ASRU)}, pages 838--844. IEEE.

\bibitem[{Peng et~al.(2018)Peng, Li, He, Liu, Bao, Wang, Song, and
  Yang}]{peng2018large}
Hao Peng, Jianxin Li, Yu~He, Yaopeng Liu, Mengjiao Bao, Lihong Wang, Yangqiu
  Song, and Qiang Yang. 2018.
\newblock Large-scale hierarchical text classification with recursively
  regularized deep graph-cnn.
\newblock In \emph{Proceedings of the 2018 world wide web conference}, pages
  1063--1072.

\bibitem[{Peng et~al.(2019)Peng, Li, Wang, Wang, Gong, Yang, Li, Yu, and
  He}]{peng2019hierarchical}
Hao Peng, Jianxin Li, Senzhang Wang, Lihong Wang, Qiran Gong, Renyu Yang,
  Bo~Li, Philip Yu, and Lifang He. 2019.
\newblock Hierarchical taxonomy-aware and attentional graph capsule rcnns for
  large-scale multi-label text classification.
\newblock \emph{IEEE Transactions on Knowledge and Data Engineering}.

\bibitem[{Qu et~al.(2020)Qu, Shen, Shen, Sajeev, Han, and Chen}]{qu2020coda}
Yanru Qu, Dinghan Shen, Yelong Shen, Sandra Sajeev, Jiawei Han, and Weizhu
  Chen. 2020.
\newblock Coda: Contrast-enhanced and diversity-promoting data augmentation for
  natural language understanding.
\newblock \emph{arXiv preprint arXiv:2010.08670}.

\bibitem[{Radford et~al.(2018)Radford, Narasimhan, Salimans, and
  Sutskever}]{radford2018improving}
Alec Radford, Karthik Narasimhan, Tim Salimans, and Ilya Sutskever. 2018.
\newblock Improving language understanding by generative pre-training.

\bibitem[{Radford et~al.(2019)Radford, Wu, Child, Luan, Amodei, and
  Sutskever}]{radford2019language}
Alec Radford, Jeffrey Wu, Rewon Child, David Luan, Dario Amodei, and Ilya
  Sutskever. 2019.
\newblock Language models are unsupervised multitask learners.
\newblock \emph{OpenAI blog}, 1(8):9.

\bibitem[{Reimers and Gurevych(2019)}]{reimers2019sentence}
Nils Reimers and Iryna Gurevych. 2019.
\newblock Sentence-bert: Sentence embeddings using siamese bert-networks.
\newblock \emph{arXiv preprint arXiv:1908.10084}.

\bibitem[{Rethmeier and Augenstein(2021)}]{rethmeier2021primer}
Nils Rethmeier and Isabelle Augenstein. 2021.
\newblock A primer on contrastive pretraining in language processing: Methods,
  lessons learned and perspectives.
\newblock \emph{arXiv preprint arXiv:2102.12982}.

\bibitem[{Veli{\v{c}}kovi{\'c} et~al.(2017)Veli{\v{c}}kovi{\'c}, Cucurull,
  Casanova, Romero, Lio, and Bengio}]{velivckovic2017graph}
Petar Veli{\v{c}}kovi{\'c}, Guillem Cucurull, Arantxa Casanova, Adriana Romero,
  Pietro Lio, and Yoshua Bengio. 2017.
\newblock Graph attention networks.
\newblock \emph{arXiv preprint arXiv:1710.10903}.

\bibitem[{Voorhees(2005)}]{voorhees2005}
Ellen Voorhees. 2005.
\newblock \href {https://doi.org/https://doi.org/10.6028/NIST.SP.500-261}
  {Overview of the trec 2004 robust retrieval track}.

\bibitem[{Wu et~al.(2020)Wu, Wang, Gu, Khabsa, Sun, and Ma}]{wu2020clear}
Zhuofeng Wu, Sinong Wang, Jiatao Gu, Madian Khabsa, Fei Sun, and Hao Ma. 2020.
\newblock Clear: Contrastive learning for sentence representation.
\newblock \emph{arXiv preprint arXiv:2012.15466}.

\bibitem[{Xiong et~al.(2020)Xiong, Xiong, Li, Tang, Liu, Bennett, Ahmed, and
  Overwijk}]{xiong2020approximate}
Lee Xiong, Chenyan Xiong, Ye~Li, Kwok-Fung Tang, Jialin Liu, Paul Bennett,
  Junaid Ahmed, and Arnold Overwijk. 2020.
\newblock Approximate nearest neighbor negative contrastive learning for dense
  text retrieval.
\newblock \emph{arXiv preprint arXiv:2007.00808}.

\bibitem[{Yang et~al.(2020)Yang, Zhang, Li, Bendersky, and
  Najork}]{yang2020beyond}
Liu Yang, Mingyang Zhang, Cheng Li, Michael Bendersky, and Marc Najork. 2020.
\newblock Beyond 512 tokens: Siamese multi-depth transformer-based hierarchical
  encoder for document matching.
\newblock \emph{arXiv preprint arXiv:2004.12297}.

\bibitem[{Yao et~al.(2019)Yao, Mao, and Luo}]{yao2019graph}
Liang Yao, Chengsheng Mao, and Yuan Luo. 2019.
\newblock Graph convolutional networks for text classification.
\newblock In \emph{Proceedings of the AAAI Conference on Artificial
  Intelligence}, volume~33, pages 7370--7377.

\bibitem[{Zaheer et~al.(2020)Zaheer, Guruganesh, Dubey, Ainslie, Alberti,
  Ontanon, Pham, Ravula, Wang, Yang et~al.}]{zaheer2020big}
Manzil Zaheer, Guru Guruganesh, Avinava Dubey, Joshua Ainslie, Chris Alberti,
  Santiago Ontanon, Philip Pham, Anirudh Ravula, Qifan Wang, Li~Yang, et~al.
  2020.
\newblock Big bird: Transformers for longer sequences.
\newblock \emph{arXiv preprint arXiv:2007.14062}.

\end{thebibliography}
\bibliographystyle{acl_natbib}

\clearpage
\appendix
\section{Training details for document classification}
We list the hyperparameters for finetuning the models on 4 document classification datasets in \cref{tab: dc_params}.

\begin{table}[h]
\centering
\scriptsize
\begin{tabular}{lcccc}
\toprule
\textbf{hyperparameters} & \multicolumn{1}{c}{\textbf{arXiv}} & \multicolumn{1}{c}{\textbf{Newsgroup}} & \multicolumn{1}{c}{\textbf{IMDB}} & \multicolumn{1}{c}{\textbf{Hyperpartisan}} \\
\midrule
learning rate & 1.00E-04 & 5.00E-05 & 5.00E-05 & 3.00E-05 \\
batch size & 32 & 32 & 32 & 32 \\
epoch & 20 & 20 & 20 & 15 \\
warmup & 0.1 & 0.1 & 0.1 & 0.1 \\
weight decay & 0.01 & 0.01 & 0.01 & 0.01 \\
\bottomrule
\end{tabular}
\caption{Hyperparameters for document classification.}
\label{tab: dc_params}
\end{table}



\end{document}